\documentclass[10pt,journal,compsoc]{IEEEtran}
\ifCLASSOPTIONcompsoc
  \usepackage[nocompress]{cite}
\else
  \usepackage{cite}
\fi
\ifCLASSINFOpdf
\else
\fi
\usepackage{multirow}
\usepackage{cite}
\usepackage{amsmath,amssymb,amsfonts}

\usepackage{graphicx}
\usepackage{algorithmic}
\def\BibTeX{{\rm B\kern-.05em{\sc i\kern-.025em b}\kern-.08em
     T\kern-.1667em\lower.7ex\hbox{E}\kern-.125emX}}

\usepackage[utf8]{inputenc}
\usepackage[english]{babel}

\usepackage[dvipsnames]{xcolor}
\begin{document}
%
\title{Bare Demo of IEEEtran.cls for\\ IEEE Computer Society Journals}
%
%
%
%
\title{Altitude Optimization of UAV Base Stations from Satellite Images Using Deep Neural Network}

 \author{\uppercase{Ibrahim Shoer}, \uppercase{Bahadir K. Gunturk}, \uppercase{Hasan F. Ates} \IEEEmembership{Senior Member, IEEE}, and \uppercase{Tuncer Baykas} \IEEEmembership{Senior Member, IEEE}


\thanks{``This work was supported by TUBITAK Grant 215E324.''}}

%
%

\markboth{}%
{Shoer \MakeLowercase{\textit{et al.}}:}
%



\IEEEtitleabstractindextext{%
\begin{abstract}
It is expected that unmanned aerial vehicles (UAVs) will play a vital role in future communication systems. Optimum positioning of UAVs, serving as base stations, can be done through extensive field measurements or ray tracing simulations when the 3D model of the region of interest is available. In this paper, we present an alternative approach to optimize UAV base station altitude for a region. The approach is based on deep learning; specifically, a 2D satellite image of the target region is input to a deep neural network to predict path loss distributions for different UAV altitudes. The predicted path distributions are used to calculate the  coverage in the region; and the optimum altitude, maximizing the coverage, is determined. The neural network is designed and trained to produce multiple path loss distributions in a single inference; thus, it is not necessary to train a separate network for each altitude. 

\end{abstract}

\begin{IEEEkeywords}
convolutional neural networks, deep learning, path loss estimation, UAV networks.
\end{IEEEkeywords}}

\maketitle

\IEEEdisplaynontitleabstractindextext

%
\IEEEpeerreviewmaketitle

\IEEEraisesectionheading{\section{Introduction}\label{sec:introduction}}
The use of unmanned aerial vehicles (UAVs) as base stations has emerged as a solution to address the demand for high-speed and reliable wireless communication in various scenarios. UAV base stations can be deployed easily, provide connectivity during emergency situations, and improve coverage and quality of service if needed.   

One of the challenging problems related to aerial base stations is the placement of the base station. The optimal base station altitude can be defined as the one that maximizes the coverage, which requires the prediction of path loss distribution in the target area. There are various models that can be used to estimate the path loss distribution, including Okumura \cite{sarkar2003survey}, Hata \cite{hata1980empirical} and Walfisch-Ikegami \cite{ChichonKurner1995} models. These models require a general classification of the area, such as "urban," "sub-urban," and "rural." Since such a rough classification of an area may not yield accurate predictions in all cases, there have been efforts to improve the results using additional information, such as the building density between transmitter and receiver \cite{5401041,6169225}. When the 3D model of an area is available, it is possible to apply ray tracing simulation techniques to get accurate results \cite{7913702, 6206329}. In addition to the requirement of a 3D model of the area, simulation based approaches have the disadvantage of high computational cost.  

There are articles specifically aiming to optimize the altitude of an aerial base station. For example, in \cite{al2014optimal}, air-to-ground channel is modeled as the addition of line-of-sight (LoS) and non-line-of-sight (NLoS) terms, where the probability of LoS depends on three parameters based on the ITU recommendation \cite{data2003prediction}: the ratio of building area to total area, the average number of buildings per unit area, and the height distribution of buildings. The formulation derived in \cite{al2014optimal} is used in other work, including \cite{alzenad20173}, where horizontal and vertical placements are decoupled, and \cite{kalantari2016number}, where the problem of placing multiple aerial base stations is addressed. In \cite{mozaffari2015drone}, using a probabilistic model for LoS, the downlink coverage performance of aerial base stations is investigated, and the optimal base station altitude, which leads to a maximum ground coverage and minimum required transmit power for a single base station, is derived. The work is extended for the placement of multiple aerial base stations in \cite{mozaffari2016efficient}. In \cite{8764728}, air-to-ground channels characteristics, including height-dependent path loss models, root mean square (RMS) delay spread and the number of multi-path components, at 3.9GHz for suburban areas and for UAV altitudes up to 40m are investigated and compared against ray tracing simulations. In \cite{8770066}, large-scale and small-scale channel parameters are extracted for LoS and NLoS cases at 1GHz and 4GHz from actual measurements; and an altitude-dependent path loss model is proposed. In \cite{7835273}, extensive measurements are taken to model air-to-ground channel characteristics, including path loss, RMS delay spread, Rician K-factor, and inter-antenna correlations for sub-urban and near-urban environments.

In recent years, the use of machine learning techniques, including deep neural networks, in wireless communication applications has become increasingly popular \cite{zhang2019deep}. Machine learning based methods have been proposed to estimate path loss \cite{Zhang2019}, classify the area model selection \cite{5624542}, identify wireless technology \cite{8292183}, identify interference \cite{schmidt2017wireless}, and allocate resources \cite{8011311}. Deep neural networks have been used to predict path loss exponent and shadowing factor \cite{ates2019} and path loss distribution \cite{Omar2020}. 

In this paper, we propose a deep learning based approach to determine path loss distribution in an area and optimize aerial base station (transmitter) altitude. Motivated by the fact that regional characteristics, such as building densities and heights, play a major role in path loss values, we aim to predict the path loss distribution of an area directly from its satellite image, which captures all regional characteristics visually. Given a dataset of satellite images and corresponding path loss distributions, a deep neural network is trained to convert an image to the corresponding path loss distribution. 

Once the path loss distributions are estimated, the coverage, i.e., the percentage of the transmitters with path loss values less than a threshold in the region, is calculated. We design our network such that it produces path loss distributions for multiple base station altitudes. Therefore, we choose the altitude which results in the maximum coverage. 

Our work is unique in the sense that it allows determining the optimal base station altitude directly from a satellite image, which is input to a deep neural network to predict path loss distribution. The path loss distribution portion of our method has similarities to the work done in \cite{Omar2020}, but there are major differences. Other than the fact that \cite{Omar2020} does not involve any altitude optimization, the network in \cite{Omar2020} produces path loss distribution for a single transmitter altitude. In this paper, the network is designed to produce multiple path loss distributions, where each distribution corresponds to a different transmitter altitude. The work in \cite{Omar2020} can be extended for multiple altitudes; however, a separate network has to be trained and stored for each transmitter altitude. In this paper, the training process involves a single network, sharing common layers for different altitudes. This reduces the overall number of parameters to be learned, which is critical in deep learning to achieve better performance for a given dataset size.



To train the proposed network, we first generate a dataset by running ray tracing simulations on 3D models of regions for which we have the satellite images as well. The ray tracing simulations produce path loss distributions, which are coupled with satellite images to form the dataset. A deep network is trained to produce path loss distributions for multiple altitudes from each satellite image. While the training process takes time, once trained, the network simply takes a satellite image to instantly produce multiple path loss distributions.

\begin{figure*}[ht]
 \centering
 \includegraphics[width=16cm,page=1]{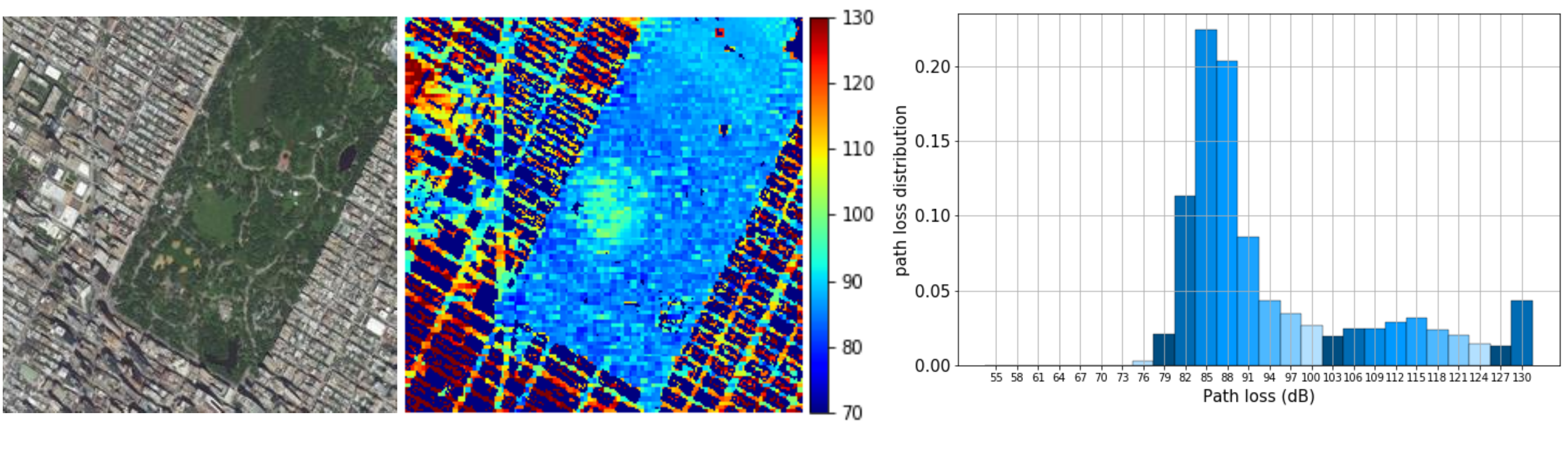}
 \caption{A satellite image of a region, the corresponding path loss map obtained through ray tracing simulations, and the path loss distribution.}
 \label{powermap}
\end{figure*}



The paper is organized as follows. In Section~\ref{sec:approach}, the proposed approach is briefly presented. The details, including the dataset generation procedure and the deep neural network architecture, are given in Section~\ref{sec:dataset} and Section~\ref{sec:prediction}, respectively. The performance of the deep neural network, the details of altitude optimization process, and the experimental results are discussed in Section~\ref{sec:results}. Finally, in Section~\ref{sec:conclusion}, the paper is concluded with a summary and possible future work.

\begin{figure*}[ht]
  \centering
 \includegraphics[width=16.4cm]{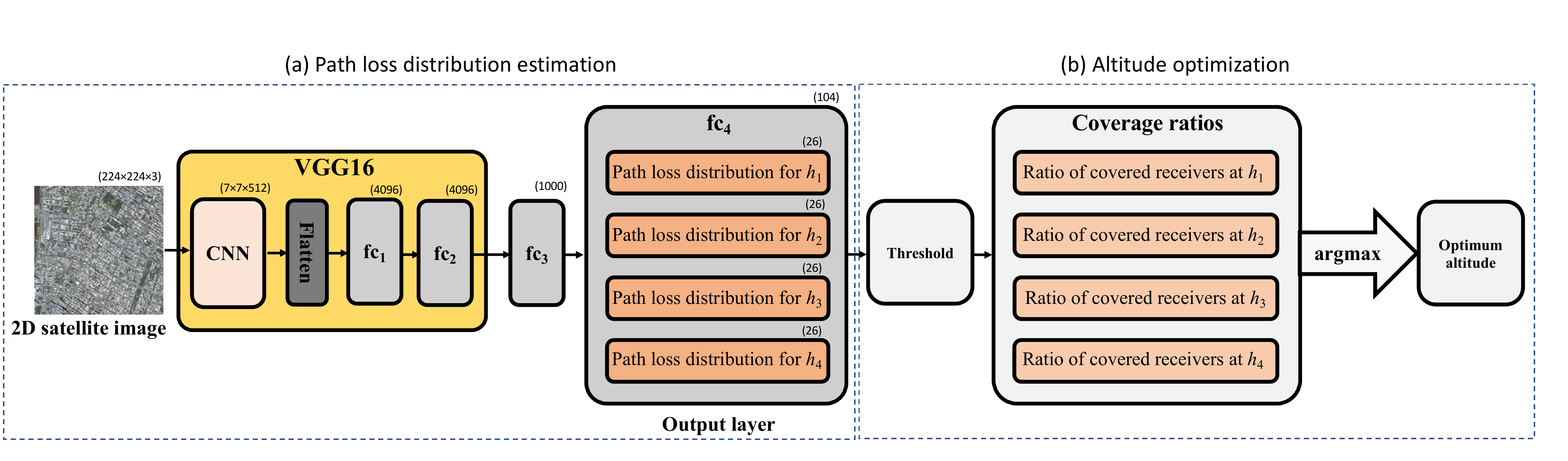} %
 \caption{(a) A satellite image of the target region is input to a deep neural network to produce path loss distributions at multiple altitudes. The network consists of a convolutional neural network (CNN) layer, followed by several fully connected layers (fc). The output layer is designed such that the path loss distribution for a specific transmitter height (${h}_i$) is produced at a specific part of the output. The initial portion of the network is taken from the VGG16 network \cite{vgg16}; the entire network is re-trained for our dataset. (b) A threshold value is chosen to determine the coverage. Receivers with path loss value greater than the threshold are not covered by the transmitter. Path loss distributions are passed through the threshold to determine the ratio of covered receivers. The altitude that has the maximum coverage ratio is the optimum altitude.}
 \label{fig:PredictionModel}
\end{figure*}

\section{Overview of the Proposed Approach}
\label{sec:approach}

The base station altitude optimization problem is defined as determining the altitude which results in the maximum number of serviceable receivers. Received signal strength, which is a function of the operation frequency, transmit power of the base station, antenna gains of the transmitter and receiver, and path loss, determines if a receiver is in the coverage area or not. Assuming there is only one transmitter and the receivers are of same type, the path loss becomes the only variable determining the coverage. The maximum allowable path loss, which we denote $PL_{th}$, can be set as the threshold; any receiver with path loss higher than this threshold is considered out of coverage. 

Fig. \ref{powermap} gives an illustration of path loss map and path loss distribution of an area. On the left of the figure, we have a satellite image of a region. The 3D model, which is not shown, of the region is input to a ray tracing software to obtain the path loss values of the receivers, which are uniformly distributed over the entire area. The path loss values are shown as a color map in the middle of the figure. The path loss values are quantized to 26 bins (with 3dB bin widths) to form the path loss distribution, which is shown on the right of the figure. The coverage (the ratio of serviceable receivers with path loss value less than a specific threshold) is calculated by summing up the path loss values less than the threshold. 

To determine the optimal altitude which yields the maximum coverage, we need to have path loss distributions corresponding to different base station altitudes. We, therefore, design our path loss estimation network to generate path loss distributions for multiple altitudes in a single inference. 

Our approach is illustrated in Fig.~\ref{fig:PredictionModel}. An image is input to a deep neural network, which consists of convolutional layers and fully connected layers. The convolutional layers extracts high and low level features, which are then flattened passed through several fully connected layers. The output of the network is a vector, which has path loss distributions for different transmitter altitudes in its designated parts. The path loss distributions are then passed through a specified threshold value to calculate coverage for each altitude. The altitude maximizing the coverage is set as the optimum altitude. 

Training a deep network requires a large dataset. In cases where the dataset size is limited, a common practice is to take a pre-trained network and re-train it with the available data for the desired application. We use this technique in our application. Specifically, we took a portion of the pre-trained VGG16 network \cite{vgg16}, appended two fully connected layers, and trained the entire network with our data. The VGG16 is pre-trained on the ImageNet dataset  \cite{imagenet}, and it extracts high and low level image features, representing the regional characteristics. The fully connected layers convert these features path loss distributions.  



 \section{Dataset Generation}
\label{sec:dataset}

The data requirement for training a deep neural network is extremely large. Generating the dataset using actual measurements were not feasible due to time, cost and legal issues of flying UAVs in urban areas. Therefore, we use the dataset generation method presented in \cite{ates2019}. Using areas with known 3D models, ray tracing simulations are used to obtain the path loss at each receiver point for each transmitter altitude. The received path loss values are then converted to path loss distributions.


Specifically, 500 geographical regions, each with a size of $1.8 \times 1.8$ km, are extracted as in \cite{ates2019}. For each region, we have the satellite image as well as the 3D model. Each 3D model is imported to a ray tracing simulation software.\footnote{https://www.remcom.com/wireless-insite-em-propagation-software} The transmitter is placed in the center of the region at a specific height. We placed 12,100 ($110 \times 110$) receivers uniformly distributed over the region at 1.5 meters above ground. We assume having a flat terrain with dry earth material and concrete buildings. We run the simulations for each transmitter altitude and each region separately. We choose 40m, 80m, 120m, and 300m as the representative set of transmitter altitudes in our experiments. The parameters used in the dataset generation are listed in Table \ref{params}.

The resulting path loss values at the receivers are processed to generate the path loss distributions. The path loss values corresponding to receivers that are inside a building are removed by utilizing the 3D models, which indicate the building locations. The remaining path loss values are quantized to 26 bins, starting from 55dB to 130dB bin centers, and bin widths of 3dB. The distributions are normalized so that the bin values add up to one for each distribution. 

We repeat the simulations for 4 different altitudes. The path loss distributions for each altitude are concatenated to construct a target vector of size 104-by-1. By repeating the simulations for different target areas, we construct a dataset of 500 images and corresponding target vectors.   


\begin{table}[ht]
\centering

\resizebox{7cm}{!}{\begin{tabular}
{|l|l|}
\hline
Transmission frequency & 900 MHz  \\ \hline
Transmission altitudes & 40, 80, 120 and 300 m \\ \hline
Transmitted power & +43 dBm \\ \hline
Receiver sensitivity & -85 dBm \\ \hline
Antenna radiation pattern & Omni-directional \\ \hline
Receiver antenna height & 1.5 m \\ \hline
Antenna polarization & Vertical \\ \hline
Transmitted signal & Sinusoid \\ \hline
Bandwidth & 8 MHz \\ \hline
\end{tabular}}
\caption{Ray tracing simulation parameters.}
\label{params}
\end{table}




\section{Network Architecture}
\label{sec:prediction}

The input to the network is of size $224 \times 224 \times 3$. The network produces 104 values, corresponding to path loss distributions for four transmitter heights.

The network consists of a common pre-trained network (VGG16~\cite{vgg16}) followed by fully connected layers. The VGG16 network is a well-known architecture used for image classification; and it is pre-trained on the ImageNet dataset  \cite{imagenet}. This portion of the network extracts the common features associated with the region. These features include low-level features such as lines, edges, corners, and other primitive features, as well as high-level features such as geometric structures and shapes with different colors and texture. Our intuition is that these features well describe the regional characteristics that directly affect the communication channel properties of the area. The last layers allow specialization/prediction of distributions for different altitudes by properly weighting the features extracted from the common layers. The final layer is a 104 unit layer with a softmax function at the end.

The dataset is divided into 400 images for training and 100 images for testing. During training, the stochastic gradient descent optimizer \cite{bottou1991stochastic} is used  with a learning rate of 0.0001, a momentum of 0.7, a batch size of 8, and cross-entropy as the loss function. 

The network can be modified (by reducing the size of the bin width and therefore increasing the number of bins) to increase the resolution of path loss distributions. This would increase the number of parameters to be learned during training; and we would need more data to successfully train the network parameters. Similarly, we can add more altitudes to the network, which would again require more training data. 

\section{Results and Discussions}
\label{sec:results}
\subsection{Deep learning model performance}

The mean squared error (MSE) between the true and predicted path loss distribution values is used to evaluate the network performance. Table \ref{tab:MSE} shows the MSE for each altitude. By comparing the MSE results to the variance of the true distribution values in the test set, it can be seen that the network is predicting the true distributions well. From the table, we notice that the prediction performance is better for higher altitudes. This improvement is due to the higher probability of LoS links for high altitudes. When the UAV is at a low altitude, such as $40$m, the communication link may suffer from shadowing due to high building, making it more difficult to predict path loss.

\begin{table}[ht]
\centering

\begin{tabular}{|c|c|c|}
\hline
Altitude & MSE [$10^{-3}$] & Test set variance [$10^{-3}$] \\ \hline
$40$ m   & 0.65    & 4.29      \\ \hline
$80$ m   & 0.32    & 2.61       \\ \hline
$120$ m   & 0.21   & 2.44      \\ \hline
$300$ m   & 0.11   & 4.42      \\ \hline
\end{tabular}
\caption{MSE between the true and predicted distribution values, and the variance of the true distribution values in the test set.}
\label{tab:MSE}
\end{table}

\begin{figure}
    \centering
    \includegraphics[page=2,scale=0.5]{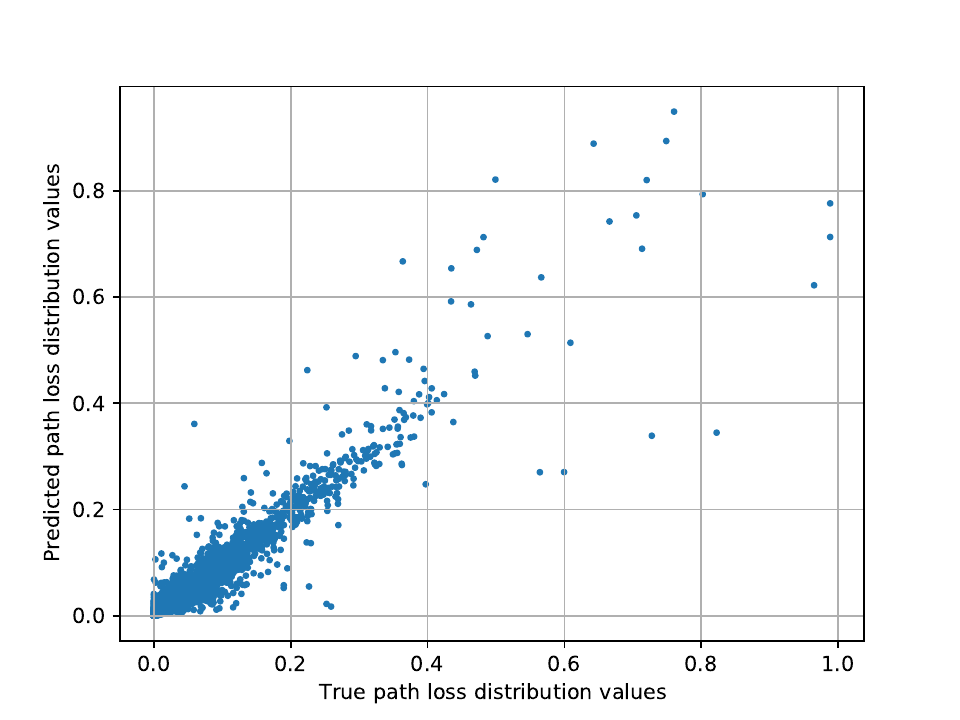}
    \caption{True versus predicted path loss distribution values.}
    \label{regression_plot}
\end{figure}

In Fig.~\ref{regression_plot}, the true versus predicted path loss distribution values are shown for the entire test set. We note that the predicted path loss values match the true values well, except for a relatively small number of outliers.

\begin{figure*} 
  \begin{tabular}{llll}

        
        & \begin{minipage}{1in}
        \includegraphics[page=13,scale=0.5,trim=0.5cm 0cm 0cm 0cm]{Figures/vgg16_output2_compressed.pdf}
      \end{minipage} & 
      \begin{minipage}{1in}
      \includegraphics[page=103,scale=0.5,trim=0.5cm 0cm 0cm 0cm]{Figures/vgg16_output2_compressed.pdf}
    \end{minipage} &  
    \begin{minipage}{1in}
      \includegraphics[page=55,scale=0.5,trim=0.5cm 0cm 0cm 0cm]{Figures/vgg16_output2_compressed.pdf}
    \end{minipage} \\
      &
    \begin{minipage}{2in}
    \includegraphics[page=14,scale=0.35]{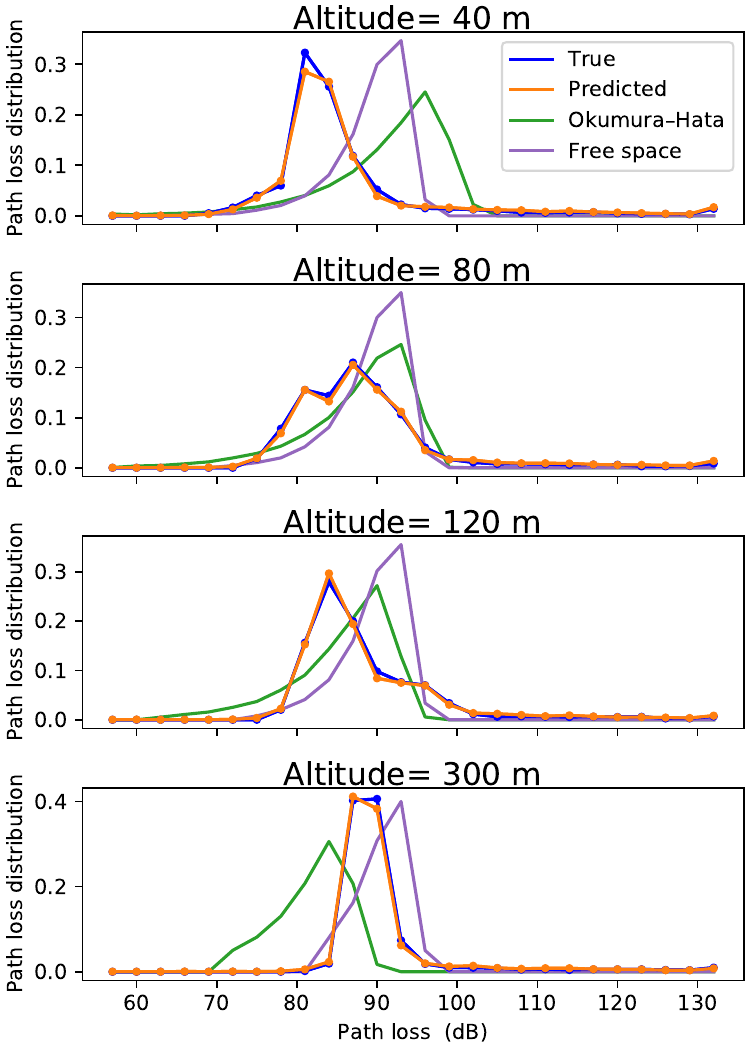}
    \end{minipage}
     &   \begin{minipage}{2in}
     \includegraphics[page=104,scale=0.35]{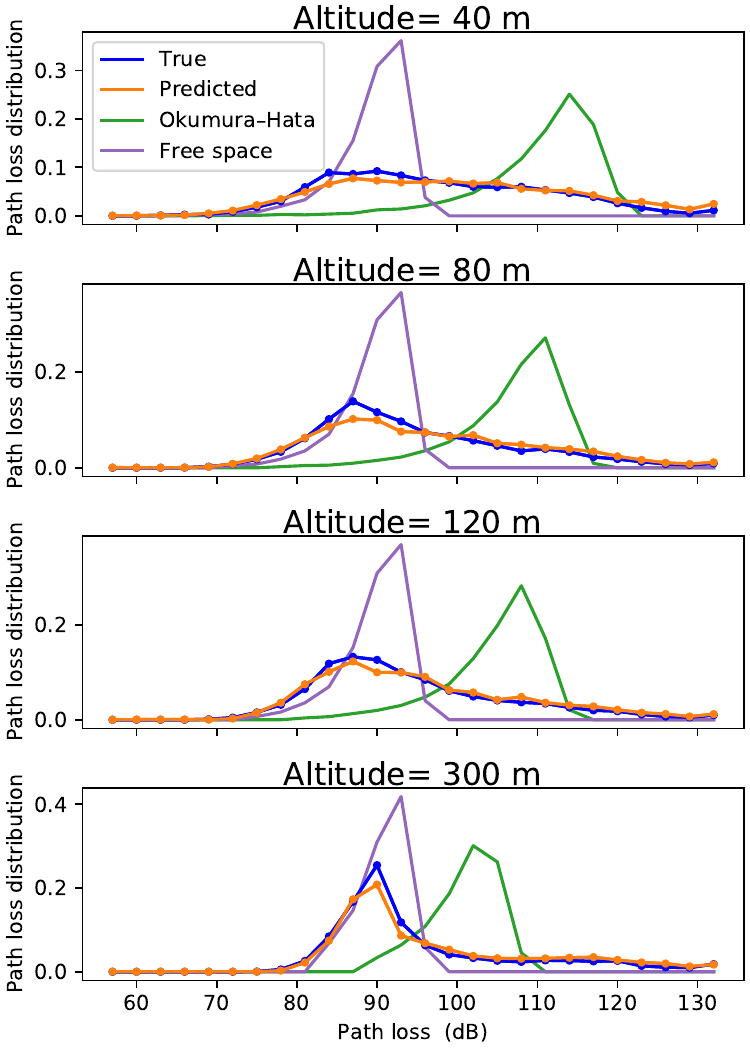}
    \end{minipage} &  \begin{minipage}{2in}
     \includegraphics[page=56,scale=0.35]{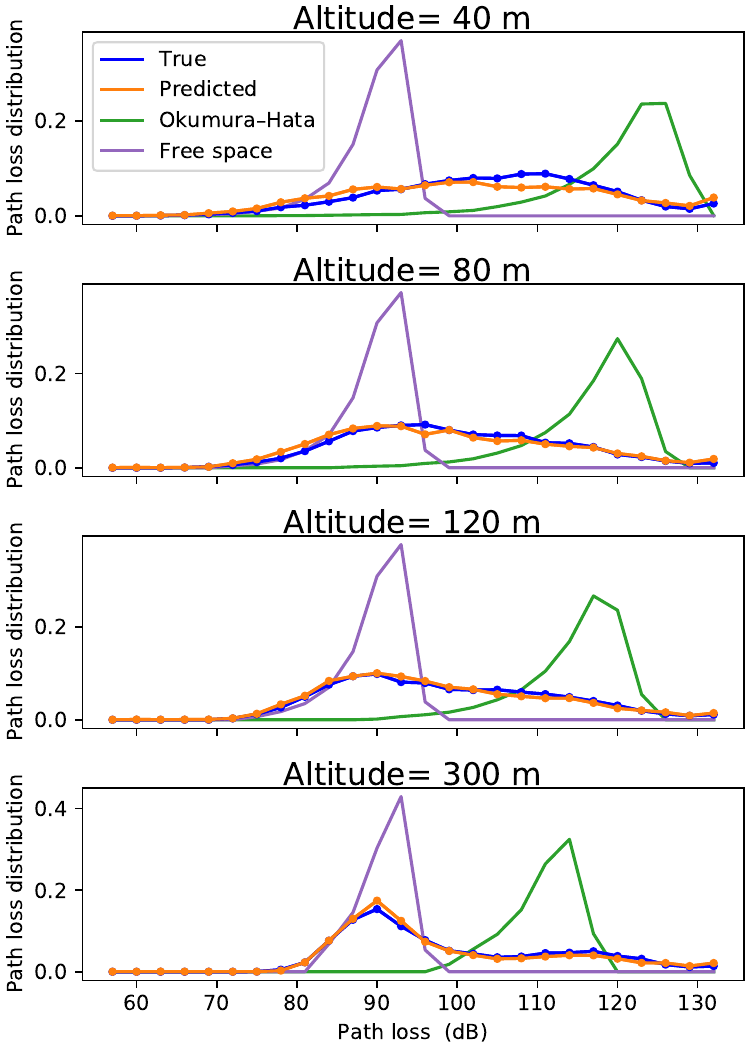}
    \end{minipage} \\  
        & \begin{minipage}{1in}
        \includegraphics[page=99,scale=0.5,trim=0.5cm 1cm 1cm 0.1cm]{Figures/vgg16_output2_compressed.pdf}
      \end{minipage} & 
      \begin{minipage}{1in}
      \includegraphics[page=159,scale=0.5,trim=0.5cm 1cm 1cm 0.1cm]{Figures/vgg16_output2_compressed.pdf}
    \end{minipage} &  
    \begin{minipage}{1in}
      \includegraphics[page=201,scale=0.5,trim=0.5cm 1cm 1cm 0.1cm]{Figures/vgg16_output2_compressed.pdf}
    \end{minipage} \\
      &
    \begin{minipage}{2in}
    \includegraphics[page=100,scale=0.35]{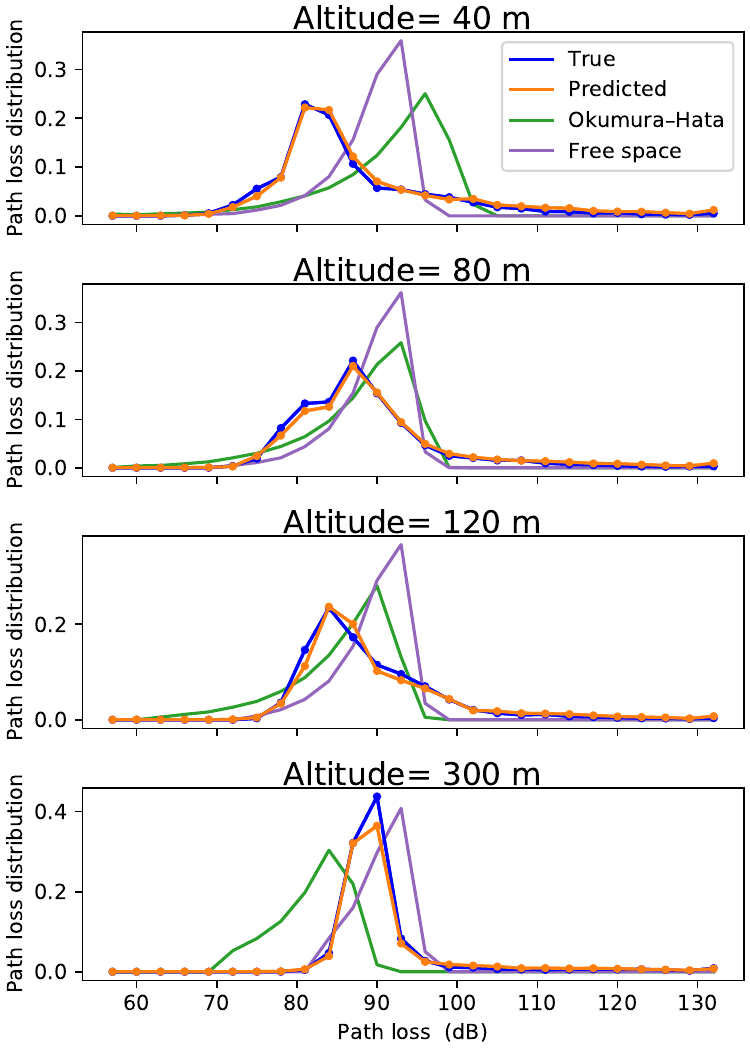}
    \end{minipage}
     &   \begin{minipage}{2in}
     \includegraphics[page=160,scale=0.35]{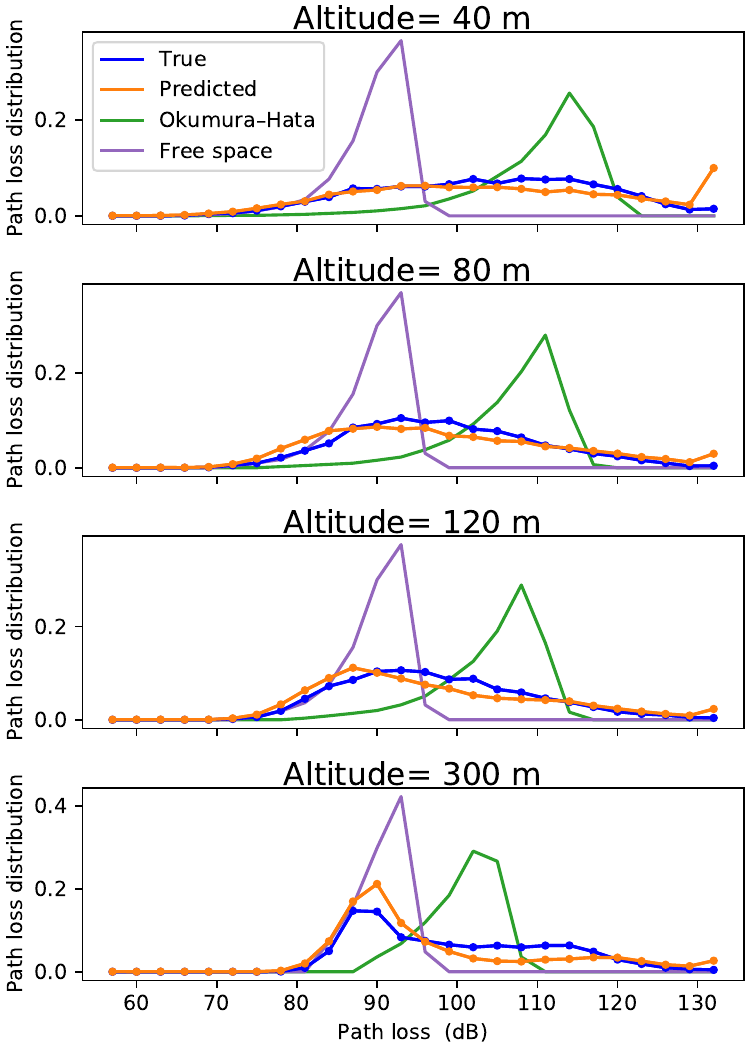}
    \end{minipage} &  \begin{minipage}{2in}
     \includegraphics[page=202,scale=0.35]{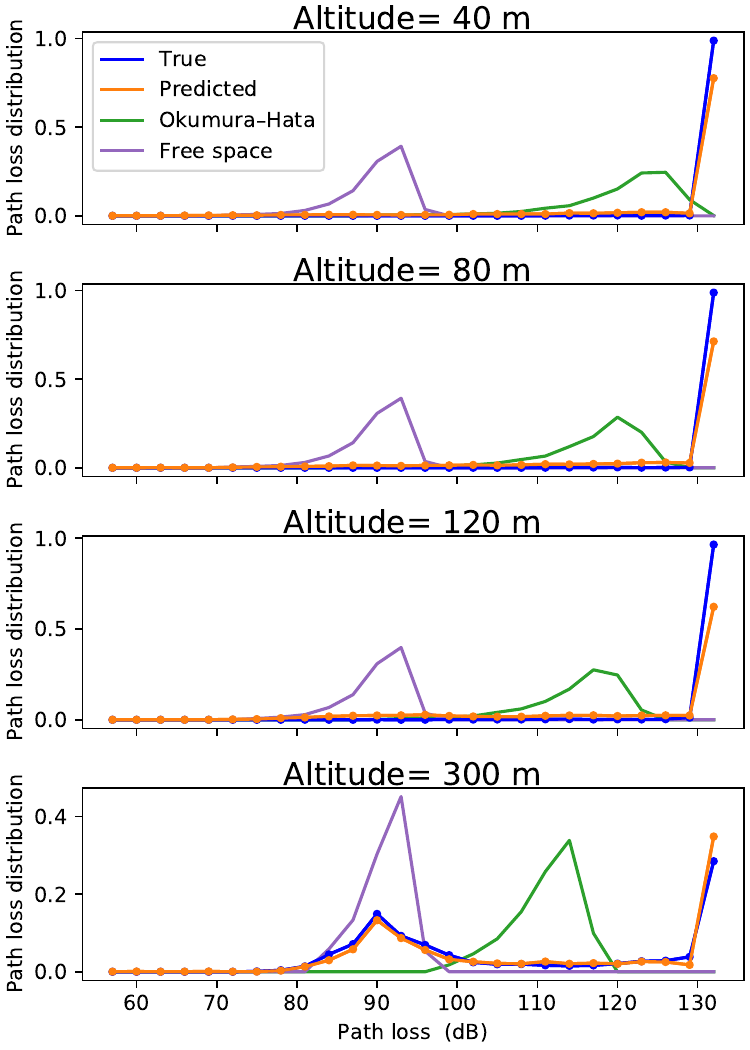}
    \end{minipage}  \\
    
  \end{tabular}
  \caption{Sample results of the proposed path loss distribution method, compared to the true distributions, and against the free-space and Okumura-Hata models for different regions and transmitter altitudes.}

    \label{tab:Output_Samples}

 \end{figure*}

In Fig.~\ref{tab:Output_Samples}, we provide some examples of the test set showing the true versus predicted distributions with the corresponding satellite images. The regions are arranged from best to worst according to the average MSE of the four altitudes; the average MSE is written above each image. We notice that regions with low MSE values correspond to sub-urban areas. The MSE increases in the regions where there are high-rise buildings, causing extreme shadowing and multi-path reflections. In the last example, there is a high-rise building in the middle of the region, blocking the transmitter signal and resulting in high path loss at almost all receiver locations. As the transmitter altitude is increased, the  receivers finally start get the signal from the transmitter. Even in such challenging situations, the network predicts the distributions well.

Fig.~\ref{tab:Output_Samples} also includes the results of two baseline methods, i.e., the free-space path loss model \cite{sarkar2003survey} and the Okumura-Hata model \cite{hata1980empirical}. The distributions predicted with the free-space path loss and Okumura-Hata models do not match well with the true distributions as these prediction methods are generic models with few categories for the regional characteristics and can not handle all sorts of variations. On the contrary, our approach can extract the regional characteristics and provide more accurate predictions. 

\begin{figure*} 
\centering
\begin{tabular}{cc}

\includegraphics[page=1,scale=0.42]{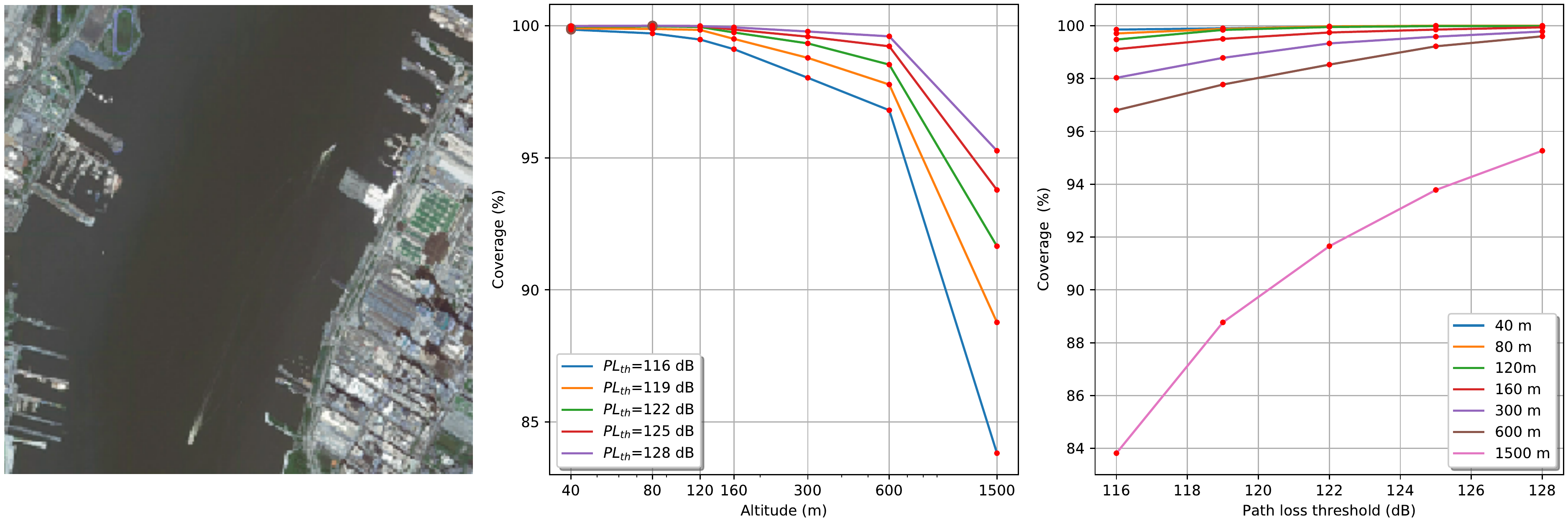} \\

\includegraphics[page=9,scale=0.42]{Figures/covplot2dtrlog.pdf} \\
\includegraphics[page=8,scale=0.42]{Figures/covplot2dtrlog.pdf}

\end{tabular}
\caption{Coverage percentages as a function of UAV altitude and path loss threshold levels for three sample regions.}

    \label{Output_Samples_true}

\end{figure*}

\subsection{Altitude optimization}

UAV altitude optimization can be done by choosing the altitude that has the maximum coverage, which is defined as the area under the path loss distribution curve below a specific threshold value. Since the distributions are quantized, the coverage is calculated by summing the values that are below the threshold value. 

To investigate altitude optimization, we performed additional simulations for different altitudes, for a few regions. Fig.~\ref{Output_Samples_true} shows the coverage curves for three different regions at seven different altitudes ($40$m, $80$m, $120$m, $160$m, $300$m, $600$m, $1500$m), and for five different path loss threshold values ($116$dB, $119$dB, $122$dB, $125$dB, $128$dB). In each row of the figure, we have the satellite image of the region, the coverage as a function of transmitter altitude, and the coverage as a function of path loss threshold. 


%

\begin{table*}
\centering

\resizebox{\textwidth}{!}{\begin{tabular}{|l|l||*{8}{c|}}
\cline{3-10}
\multicolumn{2}{c|}{}&\multicolumn{2}{c|}{\textbf{40 m}}& \multicolumn{2}{c|}{\textbf{80 m}}&\multicolumn{2}{c|}{ \textbf{120 m}}& \multicolumn{2}{c|}{\textbf{300 m}} \\ \hline
Satellite image &$PL_{th}$&True&Predicted&True&Predicted&True&Predicted&True&Predicted \\ \hline \hline
\multirow{4}{*}{\includegraphics[page=55,scale=0.14,trim=1cm 1cm 1cm 1cm,clip]{Figures/vgg16_output2_compressed.pdf}}& 128 dB 
& 0.8849 &0.9153
&0.9643&0.9701
& \textcolor{red}{\textbf{0.9657}} &\textcolor{red}{\textbf{0.9802}}  
&0.9532&0.978 \\
& 125 dB 
& 0.8614 &0.8933
& 0.9479 &0.9598
& \textcolor{red}{\textbf{0.9512}} &\textcolor{red}{\textbf{0.9713}}
& 0.9337&0.966 \\
& 122 dB 
& 0.8316 &0.864
& 0.925 &0.9429
&\textcolor{red}{\textbf{0.9308}} &\textcolor{red}{\textbf{0.9582}}
&0.9092 & 0.9487 \\
& 119 dB 
& 0.7949 &0.8294
& 0.8898 &0.9222
& \textcolor{red}{\textbf{0.9052}}&\textcolor{red}{\textbf{0.9417}}
& 0.8799& 0.9262
\\\hline
\multirow{4}{*}{\includegraphics[page=77,scale=0.14,trim=1cm 1cm 1cm 1cm,clip]{Figures/vgg16_output2_compressed.pdf}}& 128 dB
&0.962  &0.9209
& 0.9681 &\textcolor{red}{\textbf{0.9808}}
&\textcolor{red}{\textbf{0.969}} &0.9786
&0.9659 &0.9805 \\
& 125 dB
& 0.952 & 0.902
& \textcolor{red}{\textbf{0.9614}} & \textcolor{red}{\textbf{0.9722}}
& 0.9609 & 0.9705
&0.9538 & 0.9688\\
& 122 dB
& 0.9371 &0.8753
& \textcolor{red}{\textbf{0.9531}} & 0.9575
& 0.953 & \textcolor{red}{\textbf{0.9593}}
&0.9402 & 0.9502 \\
& 119 dB
& 0.9186 &0.845
& 0.9405 &0.9379
& \textcolor{red}{\textbf{0.9426}}&\textcolor{red}{\textbf{0.9431}}
& 0.9183 &0.924 
\\\hline
\multirow{4}{*}{\includegraphics[page=199,scale=0.14,trim=1cm 1cm 1cm 1cm,clip]{Figures/vgg16_output2_compressed.pdf}}& 128 dB
&0.1776 & 0.6299
&0.7695 &0.8105
&0.8787 & 0.8743
&\textcolor{red}{\textbf{0.9687}} & \textcolor{red}{\textbf{0.9988}}\\
& 125 dB
&0.131 & 0.5816
&0.7338 &0.7786
&0.8537 &0.8492
&\textcolor{red}{\textbf{0.9503}} & \textcolor{red}{\textbf{0.9827}}\\
& 122 dB
& 0.0796&0.5468
& 0.6988&0.7437
& 0.829&0.8217
& \textcolor{red}{\textbf{0.9308}}&\textcolor{red}{\textbf{0.9625}} \\
& 119 dB
& 0.0415&0.5017
&0.665 &0.7019
& 0.7968&0.7908
& \textcolor{red}{\textbf{0.9102}}&\textcolor{red}{\textbf{0.9366}} \\ 
\hline

\end{tabular}}
\caption{Coverage values calculated from true distributions and predicted distributions at different path loss thresholds $PL_{th}$ for various regions. Highest true and predicted coverage values are highlighted in "red" color.}
\label{tab:compare}
\end{table*}

In the first region, there are relatively few buildings. This means, the region can be covered well with LoS links even when the UAV altitude is low. As the altitude becomes higher, the coverage drops because of the reduction in signal strength. (See the coverage-altitude plot.) As the path loss threshold value is increased, we get more coverage as expected.  

In the second and the third regions, there are more buildings; therefore, low UAV altitudes result in less coverage. Better coverage can be achieved by increasing the UAV altitude. As the UAV altitude is increased, the coverage eventually starts to decay due to the increasing distance between the transmitter and receiver. The optimum altitude for the second region is higher than the optimum altitude for the third region because of the existence of high-rise buildings in the second region. For a path loss threshold of $128$dB, a coverage of about $95\%$ is achieved when the transmitter is at $600$m for the second region; whereas, the optimum coverage is achieved when the transmitter is at $300$m for the third region.     

We can also make further deductions by investigating the coverage-altitude and coverage-threshold plots. For example, in the first region, where we mostly have LoS links, the coverage changes little for $40$m to $600$m. The effect of altitude is more when there are buildings in the region, as in the cases of second and third images.    



Combining the path loss distribution estimations from the deep network and the altitude optimization process, we have a complete flow from an image to an optimum altitude. We tested the process with the path loss distributions obtained from the deep network. In Table \ref{tab:compare}, we exemplify coverage values obtained from true distributions and predicted distributions. We show two correct and one incorrect altitude selection. In the first and third regions, both the true distributions and predicted distributions result in the same altitude, 120m and 300m, respectively. In the second example, predicted distributions resulted in incorrect optimum altitudes for two threshold values. Even in those few incorrect cases the coverage values are close to each other at different altitudes, meaning that even an incorrect altitude is selected, similar coverage would be achieved.

\section{Conclusions}
\label{sec:conclusion}

In this paper, we present a deep learning based approach to estimate the path loss distributions in an area for multiple altitudes in a single inference from a 2D satellite image, and then use the predicted path loss distributions to optimize the altitude of a UAV, serving as a base station in an air-to-ground communication systems. 

The training process of the deep neural network is computational demanding; but once the network is trained, the inference can be done accurately in real time. This is a significant advantage over ray tracing simulations. 

While we demonstrated the idea for four different altitudes due to the extensive simulation requirements to obtain the training dataset for each altitude, the approach can be extended for more altitudes. The main challenge is the dataset generation process, which require many satellite images and ray tracing simulations on the 3D models corresponding to the satellite images. Similarly, the method can be extended for other transmission frequencies.

Another possible extension of this work is to move the transmitter position in large satellite image, crop a target region centered around the transmitter position, and predict the path loss distributions for each transmitter position using the cropped target region. In this way, both the optimum altitude and position can be determined.



\bibliographystyle{IEEEtran}
\bibliography{PL_refs}

\begin{thebibliography}{10}
\providecommand{\url}[1]{#1}
\csname url@samestyle\endcsname
\providecommand{\newblock}{\relax}
\providecommand{\bibinfo}[2]{#2}
\providecommand{\BIBentrySTDinterwordspacing}{\spaceskip=0pt\relax}
\providecommand{\BIBentryALTinterwordstretchfactor}{4}
\providecommand{\BIBentryALTinterwordspacing}{\spaceskip=\fontdimen2\font plus
\BIBentryALTinterwordstretchfactor\fontdimen3\font minus
  \fontdimen4\font\relax}
\providecommand{\BIBforeignlanguage}[2]{{%
\expandafter\ifx\csname l@#1\endcsname\relax
\typeout{** WARNING: IEEEtran.bst: No hyphenation pattern has been}%
\typeout{** loaded for the language `#1'. Using the pattern for}%
\typeout{** the default language instead.}%
\else
\language=\csname l@#1\endcsname
\fi
#2}}
\providecommand{\BIBdecl}{\relax}
\BIBdecl

\bibitem{sarkar2003survey}
T.~K. Sarkar, Z.~Ji, K.~Kim, A.~Medouri, and M.~Salazar-Palma, ``A survey of
  various propagation models for mobile communication,'' \emph{IEEE Antennas
  and propagation Magazine}, vol.~45, no.~3, pp. 51--82, 2003.

\bibitem{hata1980empirical}
M.~Hata, ``Empirical formula for propagation loss in land mobile radio
  services, ieee transa-ctions on vehicular technology,'' \emph{Vol. VT-29},
  no.~3, 1980.

\bibitem{ChichonKurner1995}
D.~J. Cichon and T.~K{\"{u}}rner, ``Propagation prediction models,''
  \emph{COST-231 TD}, vol.~95, no.~66, pp. 115--207, 1995.

\bibitem{5401041}
L.~C. Fernandes and A.~J.~M. Soares, ``Simplified characterization of the urban
  propagation environment for path loss calculation,'' \emph{IEEE Antennas and
  Wireless Propagation Letters}, vol.~9, pp. 24--27, 2010.

\bibitem{6169225}
------, ``On the use of image segmentation for propagation path loss
  prediction,'' in \emph{2011 SBMO/IEEE MTT-S International Microwave and
  Optoelectronics Conference (IMOC 2011)}, Oct 2011, pp. 129--133.

\bibitem{7913702}
B.~Ai, K.~Guan, R.~He, J.~Li, G.~Li, D.~He, Z.~Zhong, and K.~M.~S. Huq, ``On
  indoor millimeter wave massive mimo channels: Measurement and simulation,''
  \emph{IEEE Journal on Selected Areas in Communications}, vol.~35, no.~7, pp.
  1678--1690, July 2017.

\bibitem{6206329}
M.~{Zhu}, A.~{Singh}, and F.~{Tufvesson}, ``Measurement based ray launching for
  analysis of outdoor propagation,'' in \emph{2012 6th European Conference on
  Antennas and Propagation (EUCAP)}, March 2012, pp. 3332--3336.

\bibitem{al2014optimal}
A.~Al-Hourani, S.~Kandeepan, and S.~Lardner, ``Optimal lap altitude for maximum
  coverage,'' \emph{IEEE Wireless Communications Letters}, vol.~3, no.~6, pp.
  569--572, 2014.

\bibitem{data2003prediction}
P.~Data, ``Prediction methods required for the design of terrestrial broadband
  millimetric radio access systems operating in a frequency range of about
  20-50 ghz,'' \emph{Draft New Reco. ITU-R P.[DOC. 3/47], Working Party K},
  vol.~3, 2003.

\bibitem{alzenad20173}
M.~Alzenad, A.~El-Keyi, F.~Lagum, and H.~Yanikomeroglu, ``3-d placement of an
  unmanned aerial vehicle base station (uav-bs) for energy-efficient maximal
  coverage,'' \emph{IEEE Wireless Communications Letters}, vol.~6, no.~4, pp.
  434--437, 2017.

\bibitem{kalantari2016number}
E.~Kalantari, H.~Yanikomeroglu, and A.~Yongacoglu, ``On the number and 3d
  placement of drone base stations in wireless cellular networks,'' in
  \emph{2016 IEEE 84th Vehicular Technology Conference (VTC-Fall)}.\hskip 1em
  plus 0.5em minus 0.4em\relax IEEE, 2016, pp. 1--6.

\bibitem{mozaffari2015drone}
M.~Mozaffari, W.~Saad, M.~Bennis, and M.~Debbah, ``Drone small cells in the
  clouds: Design, deployment and performance analysis,'' in \emph{2015 IEEE
  Global Communications Conference (GLOBECOM)}.\hskip 1em plus 0.5em minus
  0.4em\relax IEEE, 2015, pp. 1--6.

\bibitem{mozaffari2016efficient}
------, ``Efficient deployment of multiple unmanned aerial vehicles for optimal
  wireless coverage,'' \emph{IEEE Communications Letters}, vol.~20, no.~8, pp.
  1647--1650, 2016.

\bibitem{8764728}
Z.~{Cui}, C.~{Briso}, K.~{Guan}, D.~W. {Matolak}, C.~{Calvo-Ramírez}, B.~{Ai},
  and Z.~{Zhong}, ``Low-altitude {UAV} air-ground propagation channel
  measurement and analysis in a suburban environment at 3.9{GHz},'' \emph{IET
  Microwaves, Antennas Propagation}, vol.~13, no.~9, pp. 1503--1508, 2019.

\bibitem{8770066}
Z.~{Cui}, C.~{Briso-Rodríguez}, K.~{Guan}, C.~{Calvo-Ramirez}, B.~{Ai}, and
  Z.~{Zhong}, ``Measurement-based modeling and analysis of {UAV} air-ground
  channels at 1 and 4{GHz},'' \emph{IEEE Antennas and Wireless Propagation
  Letters}, vol.~18, no.~9, pp. 1804--1808, 2019.

\bibitem{7835273}
D.~W. {Matolak} and R.~{Sun}, ``Air–ground channel characterization for
  unmanned aircraft systems—{Part III}: The suburban and near-urban
  environments,'' \emph{IEEE Transactions on Vehicular Technology}, vol.~66,
  no.~8, pp. 6607--6618, 2017.

\bibitem{zhang2019deep}
C.~Zhang, P.~Patras, and H.~Haddadi, ``Deep learning in mobile and wireless
  networking: A survey,'' \emph{IEEE Communications Surveys \& Tutorials},
  vol.~21, no.~3, pp. 2224--2287, 2019.

\bibitem{Zhang2019}
Y.~Zhang, J.~Wen, G.~Yang, Z.~He, and J.~Wang, ``Path loss prediction based on
  machine learning: Principle, method, and data expansion,'' \emph{Applied
  Sciences}, vol.~9, no. 1908, pp. 1--18, 2019.

\bibitem{5624542}
M.~Neunerdt, A.~Engels, and R.~Mathar, ``Land use classification as a key
  component for path loss prediction in rural areas,'' in \emph{2010 7th
  International Symposium on Wireless Communication Systems}, Sep. 2010, pp.
  666--670.

\bibitem{8292183}
N.~Bitar, S.~Muhammad, and H.~H. Refai, ``Wireless technology identification
  using deep convolutional neural networks,'' in \emph{2017 IEEE 28th Annual
  International Symposium on Personal, Indoor, and Mobile Radio Communications
  (PIMRC)}, Oct 2017, pp. 1--6.

\bibitem{schmidt2017wireless}
M.~Schmidt, D.~Block, and U.~Meier, ``Wireless interference identification with
  convolutional neural networks,'' in \emph{Industrial Informatics (INDIN),
  2017 IEEE 15th International Conference on}.\hskip 1em plus 0.5em minus
  0.4em\relax IEEE, 2017, pp. 180--185.

\bibitem{8011311}
U.~Challita, L.~Dong, and W.~Saad, ``Deep learning for proactive resource
  allocation in lte-u networks,'' in \emph{European Wireless 2017; 23th
  European Wireless Conference}, May 2017, pp. 1--6.

\bibitem{ates2019}
H.~F. {Ates}, S.~M. {Hashir}, T.~{Baykas}, and B.~K. {Gunturk}, ``Path loss
  exponent and shadowing factor prediction from satellite images using deep
  learning,'' \emph{IEEE Access}, vol.~7, pp. 101\,366--101\,375, 2019.

\bibitem{Omar2020}
O.~{Ahmadien}, H.~F. {Ates}, T.~{Baykas}, and B.~K. {Gunturk}, ``Predicting
  path loss distribution of an area from satellite images using deep
  learning,'' \emph{IEEE Access}, vol.~8, pp. 64\,982--64\,991, 2020.

\bibitem{vgg16}
K.~Simonyan and A.~Zisserman, ``Very deep convolutional networks for
  large-scale image recognition,'' \emph{arXiv preprint arXiv:1409.1556}, 2014.

\bibitem{imagenet}
J.~Deng, W.~Dong, R.~Socher, L.~J. Li, K.~Li, and L.~Fei-Fei, ``Imagenet: A
  large-scale hierarchical image database,'' in \emph{Proc. IEEE Conf. Comp.
  Vision Pattern Recog. (CVPR)}, 2009, pp. 248--255.

\bibitem{bottou1991stochastic}
L.~Bottou, ``Stochastic gradient learning in neural networks,''
  \emph{Proceedings of Neuro-N{\i}mes}, vol.~91, no.~8, p.~12, 1991.

\end{thebibliography}


\newpage

\begin{IEEEbiography}[{\includegraphics[width=1in,height=1.25in,clip,keepaspectratio]{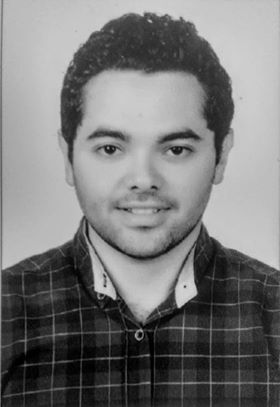}}]{Ibrahim Shoer} received his B.Sc. degree in Communications and Electronics Engineering from the Faculty of Engineering, Alexandria University, Egypt. He is pursuing his M.Sc. in Electrical Engineering and Cyber Systems, Medipol University, Istanbul, Turkey and working as a research assistant in the field of computer vision and deep learning. His current research interests include deep learning, machine learning, computer vision, signal processing, image processing and wireless communications.
\end{IEEEbiography}

\begin{IEEEbiography}[{\includegraphics[width=1in,height=1.25in,clip,keepaspectratio]{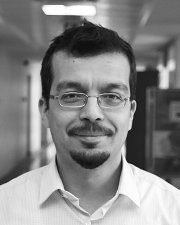}}]{Bahadir K. Gunturk} received his B.S. degree from Bilkent University, Turkey, and his Ph.D. degree from Georgia Institute of Technology in 1999 and 2003, respectively, both in electrical engineering. Between 2003 and 2014, he has been with the Department of Electrical and Computer Engineering at Louisiana State University. Since 2014, he has been with Istanbul Medipol University, where he is currently a professor. He has published more than 50 peer-reviewed journal/conference papers in the areas of image processing and computer vision.
\end{IEEEbiography}

\begin{IEEEbiography}[{\includegraphics[width=1in,height=1.25in,clip,keepaspectratio]{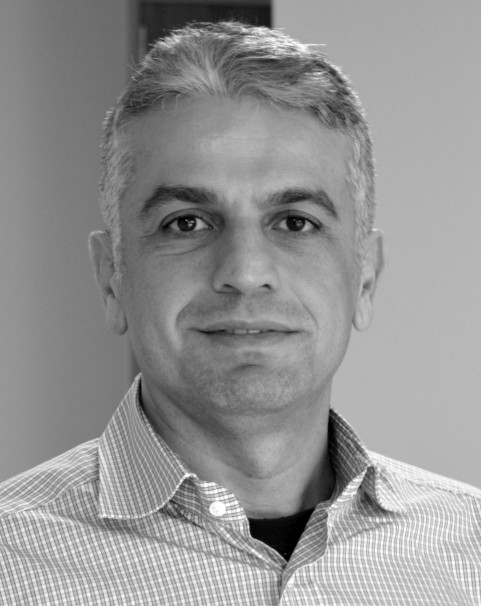}}]{Hasan F. Ates} (S'96--M'04--SM'11)  received his  PhD degree from the Department of Electrical Engineering, Princeton University in 2004. He was a Research Associate at Sabanci University, between 2004 and 2005. He held positions of Assistant, Associate and Full Professorship at Isik University between 2005 and 2018. He is currently a Professor in the Department of Computer Engineering, Istanbul Medipol University, which he joined in September 2018.  He is the author/co-author of more than 50 peer-reviewed publications in the areas of image/video processing/coding and computer vision. 
\end{IEEEbiography}

\begin{IEEEbiography}[{\includegraphics[width=1in,height=1.25in,trim= 2.8cm 2cm 2.8cm 2cm,clip]{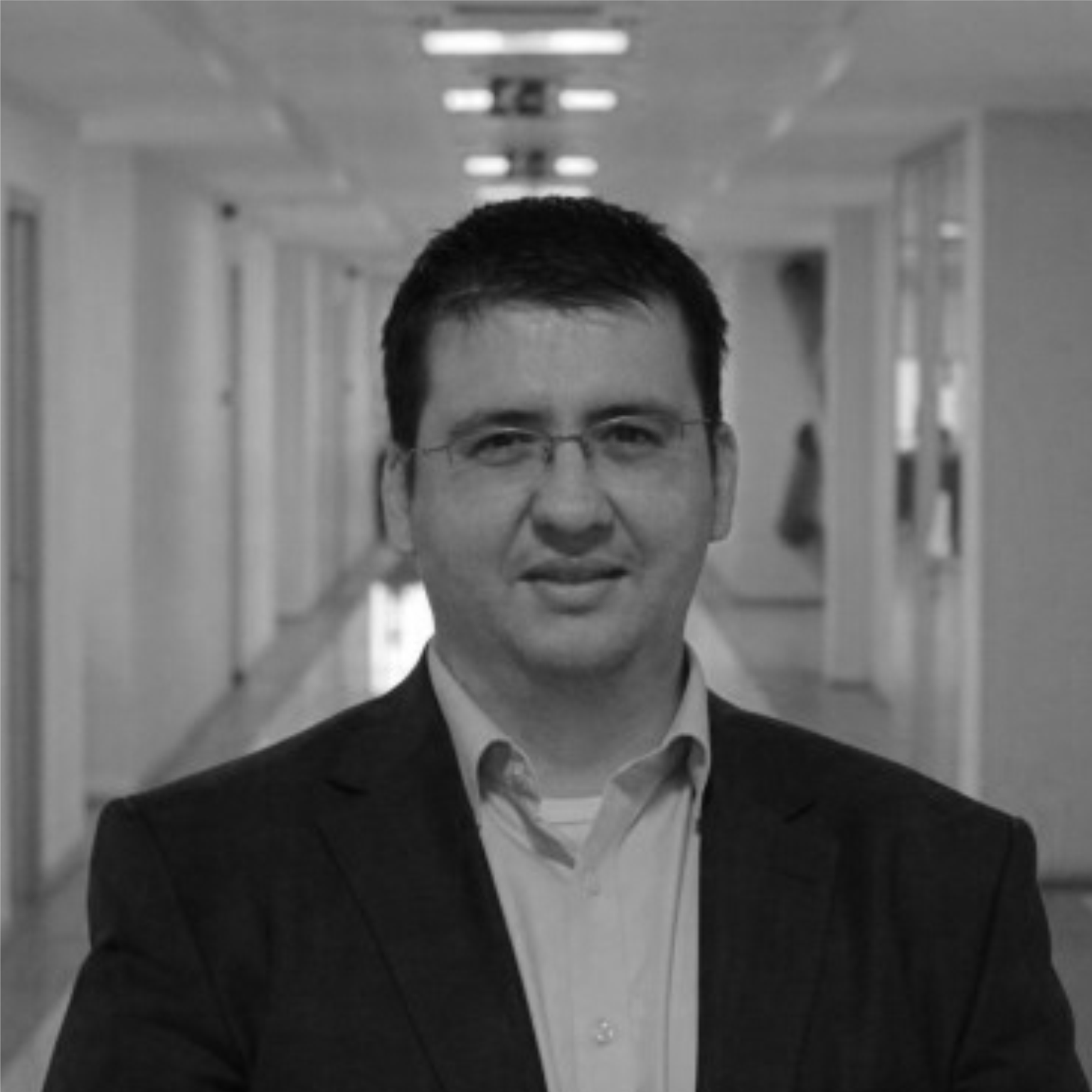}}]{Tuncer Baykas}(S'01--M'07--SM'13)  works as an assistant professor at Istanbul Medipol University. From 2007 to 2012, he worked as an expert researcher at NICT, Japan. He served as co-editor and secretary for 802.15 TG3c and contributed to multiple standardization projects, including 802.22, 802.11af and 1900.7. He is the vice chair of 802.19 Wireless Coexistence Working Group and 802.11 Task Group bb.
\end{IEEEbiography}


\end{document}